\documentclass[11pt,a4paper]{article}
\usepackage[hyperref]{acl2017}
\usepackage{times}

\usepackage{latexsym}
\usepackage{url}
\usepackage{amsmath}
\usepackage{multirow}
\usepackage{booktabs}
\usepackage{longtable}
\usepackage{enumitem}
\usepackage{graphicx}
\usepackage{color}
\usepackage{amsmath}

\usepackage{color, colortbl}
\definecolor{light-gray}{gray}{0.95}

\newcounter{example}
\newenvironment{example}[1][]{\refstepcounter{example}\par\medskip
   \noindent \textbf{Example~\theexample. #1} \rmfamily}{\medskip}

\aclfinalcopy 

\newcommand*{\affaddr}[1]{#1} 
\newcommand*{\affmark}[1][*]{\textsuperscript{#1}}

\title{SemEval 2017 Task 10: ScienceIE - Extracting Keyphrases and Relations from Scientific Publications}

\author{%
Isabelle Augenstein\affmark[1], Mrinal Das\affmark[2], Sebastian Riedel\affmark[1], Lakshmi Vikraman\affmark[2],\\ \bf and Andrew McCallum\affmark[2] \\
\affaddr{\affmark[1]Department of Computer Science, University College London (UCL), UK}\\
\affaddr{\affmark[2]College of Information and Computer Sciences, University of Massachusetts Amherst, USA} \quad \\
}

\date{}

\begin{document}
\maketitle
\begin{abstract}
We describe the SemEval task of extracting keyphrases and relations between them from scientific documents, which is crucial for understanding which publications describe which processes, tasks and materials. Although this was a new task, we had a total of 26 submissions across 3 evaluation scenarios. We expect the task and the findings reported in this paper to be relevant for researchers working on understanding scientific content, as well as the broader knowledge base population and information extraction communities.
\end{abstract}

\section{Introduction}

Empirical research requires gaining and maintaining an understanding of the body of work in specific area. For example, typical questions researchers face are which papers describe which tasks and processes, use which materials and how those relate to one another. While there are review papers for some areas, such information is generally difficult to obtain without reading a large number of publications.

Current efforts to address this gap are search engines such as Google Scholar,\footnote{\url{https://scholar.google.co.uk/}} Scopus\footnote{\url{http://www.scopus.com/}} or Semantic Scholar,\footnote{\url{https://www.semanticscholar.org/}} which mainly focus on navigating author and citations graphs.

The task tackled here is mention-level identification and classification of keyphrases, e.g. Keyphrase\_Extraction (\textsf{TASK}), as well as extracting semantic relations between keywords, e.g. Keyphrase\_Extraction \textsf{HYPONYM-OF} Information\_Extraction.
These tasks are related to the tasks of named entity recognition, named entity classification and relation extraction. However, keyphrases are much more challenging to identify than e.g. person names, since they vary significantly between domains, lack clear signifiers and contexts and can consist of many tokens. 
For this purpose, a double-annotated corpus of 500 publications with mention-level annotations was produced, consisting of scientific articles of the Computer Science, Material Sciences and Physics domains.

Extracting keyphrases and relations between them is of great interest to scientific publishers as it helps to recommend articles to readers, highlight missing citations to authors, identify potential reviewers for submissions, and analyse research trends over time. Note that organising keyphrases in terms of synonym and hypernym relations is particularly useful for search scenarios, e.g. a reader may search for articles on information extraction, and through hypernym prediction would also receive articles on named entity recognition or relation extraction.

We expect the outcomes of the task to be relevant to the wider information extraction, knowledge base population and knowledge base construction communities, as it offers a novel application domain for methods researched in that area, while still offering domain-related challenges. 

Since the dataset is annotated for three tasks dependent on one another, it could also be used as a testbed for joint learning or structured prediction approaches to information extraction~\cite{kate2010joint,singh2013joint,augenstein2015extracting,goyal2016posterior}.


Furthermore, we expect the task to be interesting for researchers studying tasks aiming at understanding scientific content, such as keyphrase extraction~\cite{kim-EtAl:2010:SemEval,hasan-ng:2014:P14-1,sterckx2016supervised,augenstein2017multitask}, semantic relation extraction~\cite{conf/lrec/TateisiSMA14,gupta-manning:2011:IJCNLP-2011,marsi-ozturk:2015:EMNLP}, topic classification of scientific articles~\cite{oseaghdha-teufel:2014:Coling}, citation context extraction~\cite{books/sp/06/Teufel06,kaplan-iida-tokunaga:2009:NLPIR4DL}, extracting author and citation graphs~\cite{peng2006information,CHAIMONGKOL14.286.L14-1259,sim-routledge-smith:2015:EMNLP} or a combination of those~\cite{radev-abujbara:2012:R50,gollapalli-li:2015:EMNLP,guo2015unsupervised}.

The expected impact of the task is an interest of the above mentioned research communities beyond the task due to the release of a new corpus, leading to novel research methods for information extraction from scientific documents. What will be particularly useful about the proposed corpus are annotations of hypernym and synonym relations on mention-level, as existing hypernym and synonym relation resources are on type-level, e.g. WordNet.\footnote{\url{https://wordnet.princeton.edu/}}
Further, we expect that these methods will directly impact industrial solutions to making sense of publications, partly due to the task organisers' collaboration with Elsevier.\footnote{\url{https://www.elsevier.com/}}

\section{Task Description}\label{sec:task_description}

The task is divided into three subtasks:
\begin{enumerate}
\item[A)]{Mention-level keyphrase identification}
\item[B)]{Mention-level keyphrase classification. Keyphrase types are \textsf{PROCESS} (including methods, equipment), \textsf{TASK} and \textsf{MATERIAL} (including corpora, physical materials)}
\item[C)]{Mention-level semantic relation extraction between keyphrases with the same keyphrase types. Relation types used are \textsf{HYPONYM-OF} and \textsf{SYNONYM-OF}.}
\end{enumerate}
We will refer to the above subtasks as Subtask A, Subtask B, and Subtask C respectively. 

A shortened (artificial) example of a data instance for the Computer Science area is displayed in Example~\ref{ex:Short}, examples for Material Science and Physics are included in the appendix.
The first part is the plain text paragraph (with keyphrases in italics for better readability), followed by stand-off keyphrase annotations based on character offsets, followed relation annotations.

\newpage

\begin{example}\label{ex:Short}
\\\textbf{Text}: \textit{Information extraction} is the process of extracting structured data from unstructured text, which is relevant for several end-to-end tasks, including \textit{question answering}. This paper addresses the tasks of \textit{named entity recognition} (\textit{NER}), a subtask of \textit{information extraction}, using \textit{conditional random fields} (\textit{CRF}). Our method is evaluated on the \textit{ConLL-2003 NER corpus}. \\
\begin{table}[h]
\centering
\small
\begin{tabular}{|llll|}
\hline
\rowcolor{light-gray} 
\bf ID & \bf Type & \bf Start & \bf End \\
\hline
0 & TASK & 0 & 22 \\
1 & TASK & 150 & 168 \\
2 & TASK & 204 & 228 \\
3 & TASK & 230 & 233 \\
4 & TASK & 249 & 271 \\
5 & PROCESS & 279 & 304 \\
6 & PROCESS & 306 & 309 \\
7 & MATERIAL & 343 & 364 \\
\hline
\end{tabular}
\\
\vspace{0.1in}
\begin{tabular}{|lll|}
\hline
\rowcolor{light-gray} 
\bf ID1 & \bf ID2 & \bf Type \\
\hline
2 & 0 & HYPONYM-OF \\
2 & 3 & SYNONYM-OF \\
5 & 6 & SYNONYM-OF \\
\hline
\end{tabular}
\end{table}
\end{example}

\begin{figure}[h]
\centering
\includegraphics[width=\linewidth]{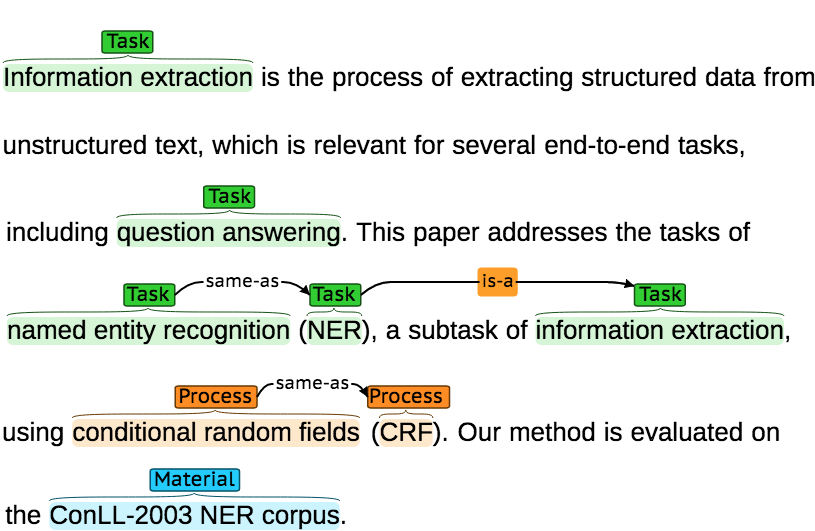}
\end{figure}

\section{Resources for SemEval-2017 Task}

\subsection{Corpus}

A corpus for the task was built from ScienceDirect\footnote{\url{http://www.sciencedirect.com/}} open access publications and was available freely for participants, without the need to sign a copyright agreement. Each data instance consists of one paragraph of text, drawn from a scientific paper.

Publications were provided in plain text, in addition to xml format, which included the full text of the publication as well as additional metadata. 500 paragraphs from journal articles evenly distributed among the domains Computer Science, Material Sciences and Physics were selected.

The training data part of the corpus consists of 350 documents, 50 for development and 100 for testing. This is similar to the pilot task described in Section~\ref{sec:Pilot}, for which 144 articles were used for training, 40 for development and for 100 testing.

We present statistics about the dataset in Table~\ref{tab:DataStatistics}. Notably, the dataset contains many long keyphrases. 22\% of all keyphrases in the training set consist of words of 5 or more tokens. This contributes to making the task of keyphrase identification very challenging. However, 93\% of those keyphrases are noun phrases\footnote{Parts of speech are determined automatically, using the nltk POS tagger}, which is valuable information for simple heuristics to identify keyphrase candidates. Lastly, 31\% of keyphrases contained in the training dataset only appear in it once, systems will have do generalise to unseen keyphrases well. 

\setlength{\tabcolsep}{0.3em}
\begin{table*}[t]
\fontsize{10}{10}\selectfont
\begin{center}
\begin{tabular}{|l | l|}
\hline
\rowcolor{light-gray} 
Characteristic & \\
\hline
Labels & Material, Process, Task \\
Topics & Computer Science, Physics, Material Science  \\
Number all keyphrases & 5730 \\
Number unique keyphrases & 1697 \\
\% singleton keyphrases & 31\% \\
\% single-word mentions & 18\% \\
\% mentions, word length $>=$ 3 & 51\% \\
\% mentions, word length $>=$ 5 & 22\% \\
\% mentions, noun phrases & 93\% \\
Most common keyphrases & `Isogeometric analysis', `samples', `calibration process',`Zirconium alloys'\\
\hline
\end{tabular}
\end{center}
\caption{\label{tab:DataStatistics} Characteristics of SemEval 2017 Task 10 dataset, statistics of training sets}
\end{table*}

\subsection{Annotation Process}

Mention-level annotation is very time-consuming, and only a handful of semantic relations such as hypernymy and synonymy can be found in each publication. We therefore only annotate paragraphs of publications likely to contain relations.

We originally intended to identify suitable documents by automatically extracting a knowledge graph of relations from a large scientific dataset using Hearst-style patterns~\cite{hearst1991noun,snow2005}, then using those to find potential relations in a distinct set of documents, similar to the distant supervision~\cite{mintz2009distant,snow2005} heuristic. Documents containing a high number of such potential relations would then be selected.
However, this requires automatically learning to identify keyphrases between which those potential relations hold, and requires relations to appear several times in a dataset for such a knowledge graph to be useful.

In the end, this strategy was not feasible due to the difficulty of learning to detect keyphrases automatically and only a small overlap between relations in different documents. Instead, keyphrase-dense paragraphs were detected automatically using a coarse unsupervised approach~\cite{NIPS2013_5021} and those likely to contain relations were selected manually for annotation.

For annotation, undergraduate student volunteers studying Computer Science, Material Science or Physics were recruited using UCL's student newsletter, which reaches all of its students. Students were shown example annotations and the annotation guidelines, and if they were still interested in participating in the annotation exercise, afterwards asked to select beforehand how many documents they wanted to annotate. Approximately 50\% of students were still interested, having seen annotated documents and read annotation guidelines. They were then given two weeks to annotate documents with the BRAT tool~\cite{stenetorp2012brat}, which was hosted on an Amazon EC2 instance as a web service.
Students were compensated for annotations per document. Annotation time was estimated as approximately 12 minutes per document and annotator, on which basis they were paid roughly 10 GBP per hour. They were only compensated upon completion of all annotations, i.e. compensation was conditioned on annotating all documents. The annotation cost was covered by Elsevier.
To develop annotation guidelines, a small pilot annotation exercise on 20 documents was performed with one annotator after which annotation guidelines were refined.\footnote{Annotation guidelines were available to task participants, they can be found here: \url{https://scienceie.github.io/resources.html}}

We originally intended for student annotators to triple annotate documents and apply majority voting on the annotations, but due to difficulties with recruiting high-quality annotators we instead opted to double-annotate documents, where the second annotator was an expert annotator. Where annotations disagreed, we opted for the expert's annotation. Pairwise inter-annotator agreement between the student annotator and the expert annotator measured with Cohen's kappa is shown in Table~\ref{tab:iaa}. The * indicates annotation quality decreased over time, ending with the annotator not completing annotating all documents. To account for this, documents for which no annotations are given are excluded from computing inter-annotator agreement. Out of the annotators completing the annotation exercise, Cohen's kappa ranges between 0.45 and 0.85, with half of them having a substantial agreement of 0.6 or higher. For future iterations of this task, we recommend to invest significant efforts into recruiting high-quality annotators, perhaps with more pre-annotation quality screening.

\begin{table}[t]
\centering
\begin{tabular}{|l|c|c|c|}
\hline
\rowcolor{light-gray} 
Student Annotator &	IAA \\
\hline
1 & 0.85 \\
2 & 0.66 \\
3 & 0.63 \\
4 & 0.60 \\
\hline
5 & 0.50 \\
6 & 0.48 \\
7 & 0.47 \\
8 & 0.45 \\ 
\hline
9* & 0.25 \\
10* & 0.22 \\
11* & 0.20 \\
12* & 0.15 \\
13* & 0.06 \\
\hline
\end{tabular} \caption{Inter-annotator agreement between the student annotator and the expert annotator, measured with Cohen's Kappa}
\label{tab:iaa}
\end{table}


\section{Evaluation}\label{sec:Evaluation}

SemEval 2017 Task 10 offers three different evaluation scenarios:
\begin{itemize}
\item[1)]{Only plain text is given (Subtasks A, B, C).}
\item[2)]{Plain text with manually annotated keyphrase boundaries are given (Subtasks B, C).}
\item[3)]{Plain text with manually annotated keyphrases and their types are given (Subtask C).}
\end{itemize}
We refer to the above scenarios as Scenario 1, Scenario 2, and Scenario 3 respectively. 

\subsection{Metrics}
Keyphrase identification (Subtask A) has traditionally been evaluated by calculating the exact matches with the gold standard. There is existing work for capturing semantically similar keyphrases ~\cite{zesch2009approximate,kim2010evaluating}, however since these are captured using relations, similar to the pilot task on keyphrase extraction (Section~\ref{sec:Pilot}) we evaluate keyphrases, keyphrase types and relations with exact match criteria. The output of systems is matched exactly against the gold standard. The traditionally used metrics of precision, recall and F1-score are computed and the micro-average of those metrics across publications of the three genres are calculated. These metrics are also calculated for Subtasks B and C. In addition, for Subtasks B and C, participants are given the option of using text manually annotated with keyphrase mentions and types.

\section{Pilot Task}\label{sec:Pilot}

A pilot task on keyphrase extraction from scientific documents was run by other organisers at SemEval 2010~\cite{kim-EtAl:2010:SemEval}. The task was to extract a list of keyphrases representing key topics from scientific documents, i.e. similar to the first part of our proposed Subtask A, only on type-level. Participants were allowed to submit up to 3 runs and were required to submit a list of 15 keyphrases for each document, ranked by the probability of being reader-assigned phrases.
Data was collected from the ACM Digital Library for the research areas Distributed Systems, Information Search and Retrieval, Distributed Artificial Intelligence – Multiagent Systems and Social
and Behavioral Sciences – Economics. Participants were provided with 144 training, 40 development and 100 test articles, each set containing a mix of articles of the different research areas. The data was provided in plain text, converted from pdf with \textsf{pdftotext}. Publications were annotated with keyphrases by 50 Computer Science students and added to author-provided keyphrases required by the journals they were published in. Guidelines were for the keyphrases to exactly appear anywhere in the text of the paper, in reality 15\% of annotator-provided keyphrases did not, as well as 19\% of author-provided keyphrases. The number of author-specified keywords was 4 on average, whereas annotators identified 12 on average.
Returned phrases are considered correct if they are exact matches of either the annotator- or author-assigned keyphrases, allowing for minor syntactic variations (A of B $\rightarrow$ B A ; A's
B $\rightarrow$ A B). Precision, recall and F1 is calculated for the top 5, top 10 and all keywords.
19 systems were submitted to the task, the best one achieving an F1 of 27.5\% on the combined author-assigned and annotator-assigned keywords.

Lessons learned from the task were that performance varies depending on how many keywords are to be extracted, the task organisers recommend against fixing a threshold for a number of keyphrases to extract lead. They further recommend a more semantically-motivated task, taking into account synonyms of keyphrases instead of requiring exact matches. Both of those recommendations will be taken into account for future task design. To fulfill the latter, we will ask annotator to assign types to the identified keywords (process, task, material) and identify semantic relations between them (hypernym, synonym). 

\section{Existing Resources}\label{sec:Existing}

As part of the FUSE project with IARPA, we created a small annotated corpus of 100 noun phrases generated from the titles and abstracts derived from the Web Of Science corpora\footnote{\url{http://thomsonreuters.com/en/products-services/scholarly-scientific-research/scholarly-search-and-discovery/web-of-science.html}} of the domains Physics, Computer Science, Chemistry and Computer Science. These corpora cannot be distributed publicly and were made available by the IARPA funding agency. Annotation was performed by 3 annotators using 14 fine-grained types, including \textsf{PROCESS}.

We measured inter-annotator agreement among the three annotators for the 14 categories using Fleiss' Kappa. The k value was found to be 0.28 which implies that there was fair agreement between them, however distinguishing between the fine-grained types added significantly to the annotation time. Therefore we only use three main types for the SemEval 2017 Task 10. 

There are some existing keyphrase extraction corpora, however, they are not similar enough to the proposed task to justify reuse. Below is a description of existing corpora.

The SemEval 2010 Keyphrase Extraction corpus~\cite{kim-EtAl:2010:SemEval}\footnote{\url{https://github.com/snkim/AutomaticKeyphraseExtraction}} consists of a handful of document-level keyphrases per article. In contrast to the task proposed, the keyphrases are annotated on type-level and not further classified as process, task or material and semantic relations are not annotated. Further, the domains considered are different and mostly sub-domains of Computer Science. 

The corpus released by~\newcite{conf/lrec/TateisiSMA14}\footnote{\url{https://github.com/mynlp/ranis}} contains sentence-level fine-grained semantic annotations for 230 publication abstracts in Japanese and 400 in English. In contrast to what we propose, the annotations are more fine-grained and annotations are only available for abstracts.

\newcite{gupta-manning:2011:IJCNLP-2011} studied keyphrase extraction from ACL Anthology articles, applying a pattern-based bootstrapping approach based on 15 016 documents and assigning the types \textsf{FOCUS}, \textsf{TECHNIQUE} and \textsf{DOMAIN}. Performance was evaluated on 30 manually annotated documents.
Although the latter corpus is related to what we propose, manual annotation is only available for a small number of documents and only for the Natural Language Processing domain.

The ACL RD-TEC 2.0 dataset \cite{conf/lrec/QasemiZadehS16} consists of 300 ACL Anthology abstracts annotated on mention-level with seven different types of keyphrases. Unlike our dataset, it does not contain relation annotations. Note that this corpus was created at the same time as the one SemEval 2017 Task 10 dataset and thus we did not have the chance to build on it. A more in-depth comparison between the two datasets as well as keyphrase identification and classification methods evaluated on them can be found in~\newcite{augenstein2017multitask}.

\subsection{Baselines}

We frame the task as a sequence-to-sequence prediction task.
We preprocess the files by splitting documents into sentences and tokenising them with nltk, then aligning span annotations from .ann files to tokens. Each sentence is regarded as one sequence.
We then split the task into the three subtasks, keyphrase boundary identification, keyphrase classification and relation classification and add three output layers. We predict the following types, for the three subtasks respectively:\\
Subtask A: $t_A = {O, B, I}$ for tokens being outside, at the beginning, or inside a keyphrase\\
Subtask B: $t_B = {O, M, P, T}$ for tokens being outside a keyphrase, or being part of a material, process or task\\
Subtask C: $t_C = {O, S, H}$ for Synonym-of and Hyponym-of relations.
For Subtask A and B, we predict one output label per input token. For Subtask C we predict a vector for each token, that encodes what the relationship between that token and every other token in the sequence is for the first token in each keyphrase.
After predictions for tokens are obtained, these are converted back to spans and relations between them in a post-processing step.

We report results for two simple models: one to estimate the \textit{upper bound}, that converts .ann files into instances, as described above, then converts them back into .ann files.
Next, to estimate a lower bound, a \textit{random baseline}, that for each token assigns a random label for each of the subtasks.

The \textit{upper bound} span-token-span round-trip conversion performance, an F1 of 0.84, shows that we already lose a significant amount of performance due to sentence splitting and tokenisation alone. The \textit{random} baseline further shows hard especially the keyphrase boundary identification task is and as a result the overall task, since the subtasks depend on one another. For Subtask A, a random baseline achieves an F1 of 0.03. The overall tasks gets easier if keyphrase boundaries are given, resulting in F1 of 0.23 for keyphrase classification, and if keyphrase types are given, an F1 of 0.04 are achieved with the random baseline for Subtask C.


\section{Summary of Participating Systems}


\begin{table*}[t]
\centering
\begin{tabular}{|l|c|c|c|c|}
\hline
\rowcolor{light-gray} 
Teams      &	Overall &	A &	B	& C \\
\hline
s2\_end2end \cite{ammar2017scienceie}      &	{\bf 0.43} & 0.55	& {\bf 0.44} & {\bf 0.28} \\
TIAL\_UW	        &   0.42 & {\bf 0.56}	& {\bf 0.44} &	 \\
TTI\_COIN \cite{tomoki2017scienceie}        &	0.38 & 0.5	& 0.39 & 0.21 \\
PKU\_ICL \cite{wang2017scienceie}	        &   0.37 & 0.51	& 0.38 & 0.19 \\
NTNU-1 \cite{marsi2017scienceie}	&   0.33 & 0.47	& 0.34 & 0.2 \\
WING-NUS \cite{prasad2017scienceie}        &	0.27 & 0.46	& 0.33 & 0.04 \\
Know-Center \cite{kern2017scienceie}     &	0.27 & 0.39	& 0.28 &	\\
SZTE-NLP \cite{berend2017scienceie}        &	0.26 & 0.35	& 0.28 &	\\
NTNU \cite{lee2017scienceie}           &	0.23 & 0.3	& 0.24 & 0.08 \\
LABDA \cite{segurabedmar2017scienceie}           &	0.23 & 0.33	& 0.23 &	\\
LIPN \cite{hernandez2017scienceie}            &	0.21 & 0.38	& 0.21 & 0.05 \\
SciX            &	0.2	 & 0.42	& 0.21 &	\\
IHS-RD-BELARUS  &	0.19 & 0.41	& 0.19 &	\\
HCC-NLP	        &   0.16 & 0.24	& 0.16 &	\\
NITK\_IT\_PG	    &   0.14 & 0.3	& 0.15 &	\\
Surukam	        &   0.1  & 0.24	& 0.1  & 0.13 \\
GMBUAP \cite{flores2017scienceie}	        &   0.04 & 0.08	& 0.04 &	\\
\hline
\textit{upper bound} &   0.84 &  0.85 &  0.85 & 0.77	\\
\textit{random} &  0.00  & 0.03  & 0.01  & 0.00	\\
\hline
\end{tabular} \caption{F1 scores of teams participating in Scenario 1 and baseline models for Overall, Subtask A, Subtask B, and Subtask C. Ranking of the teams is based on overall performance measured in Micro F1. } 
\label{tab:scenario1}
\end{table*}

In this section, we summarise the outcome of the competition. For more details please refer to the respective system description papers and the task website \url{https://scienceie.github.io/}.

We had three subtasks, described in Sec \ref{sec:task_description}, which were grouped together in three evaluation scenarios, described in Sec \ref{sec:Evaluation}. 
The competition was hosted in CodaLab\footnote{\url{https://competitions.codalab.org/competitions/15898}} in two phases: (i) development phase and (ii) testing phase. Fifty four teams participated in the development phase, and out of them twenty six teams participated in the final competition. One of the major success of the competition is due to such wide participation and application of various different techniques starting from neural networks, supervised classification with careful feature engineering to simple rule based methods. We present a summary of approaches used by task participants below.

\subsection{Evaluation Scenario 1}
In this scenario teams need to solve all three sub-tasks A, B, and C; where no annotation information was given. Some teams participated only in Subtask A, or B; but the overall micro F1 performance across subtasks is considered for the ranking of the teams. Seventeen teams participated in this scenario. The F1 scores range from 0.04 to 0.43. Complete results are given in Table \ref{tab:scenario1}.

Various different types of methods have been applied by different teams with various levels of supervision. The best three teams TTI\_COIN, TIAL\_UW, and s2\_end2end  have used recurrent neural network (RNN) based approaches to obtain F1 scores of 0.38, 0.42 and 0.43 respectively. However, TIAL\_UW, and s2\_end2end, by using a conditional random fields (CRF) layer on top of RNNs achieve a higher F1 in Subtask A compared to TTI\_COIN. 

The fourth team PKU\_ICL with an F1 of 0.37 found classification models based on random forest and support vector machines (SVM) useful with carefully engineered feature such as TF-IDF over a very large external corpus, IDF weighted word-embeddings etc, along with an existing taxonomy. 
SciX on the other hand used noun phrase chunking and trained an SVM classifier on provided training data to classify phrases, and used a CRF to predict labels of the phrases. 
CRF based methods with parts-of-speech (POS) tagging and orthographic features such as presence of symbols and capitalisation have been tried by several teams (NTNU, SZTE-NLP, WING-NUS) and they leading to a reasonable performance (F1: 0.23, 0.26, and 0.27, respectively).

Noun phrase extraction with length constraint by HCC-NLP, and using a global list of keyphrases by NITK\_IT\_PG are found not to perform satisfactorily (F1: 0.16 and 0.14 respectively). The former is surprising, as keyphrases are with an overwhelming majority noun phrases, the latter not as much, many keyphrases only appear once in the dataset (see Table~\ref{tab:DataStatistics}).
GMBUAP further tried using empirical rules obtained by observing the training data for Subtask A, and a Naive Bayes classifier trained on provided training data for Subtask B. Such simple methods on their own prove not to be accurate enough. Attempts of such give us additional insight about the hardness of the problem and applicability of simple methods to the task. 

\begin{table*}[t]
\centering
\begin{tabular}{|l|c|c|c|}
\hline
\rowcolor{light-gray} 
Teams      &	Overall &	B	& C \\
\hline
MayoNLP \cite{liu2017scienceie}	    & {\bf 0.64} & {\bf 0.67} &	{\bf 0.23} \\
UKP/EELECTION \cite{eger2017scienceie}\footnotemark	        & 0.63 & 0.66 &	\\
LABDA \cite{segurabedmar2017scienceie}       & 0.48 & 0.51 &	\\
BUAP \cite{aleman2017scienceie}        & 0.43 & 0.45 & \\
\hline
\textit{upper bound} &   0.84 &  0.85 & 0.77	\\
\textit{random} & 0.15 & 0.23 & 0.01 \\
\hline
\end{tabular} \caption{F1 scores of teams participating in Scenario 2 and baseline models for Overall, Subtask B, and Subtask C. Ranking of the teams is based on overall performance measured in Micro F1. Teams participating in Scenario 2 received partial annotation with respect to Subtask A.} 
\label{tab:scenario2}
\end{table*}

\footnotetext{After the end of the evaluation period, team UKP/EELECTION discovered those results were based on training on the development set. For training on the training set, their results are: 0.69 F1 overall and 0.72 F1 for Subtask B only}
\addtocounter{footnote}{1}

\subsection{Evaluation Scenario 2}
In this scenario teams needed to solve sub-tasks B, and C. Partial annotation was provided to the teams, that is, solution to the Subtask A. 
Four teams participated in this scenario with F1 cores ranging from 0.43 to 0.64. Please refer to Table \ref{tab:scenario2} for complete result.

Except MayoNLP, other three teams participated only in Subtask B. Although ranking is done based on overall performance, but in this scenario rankings are consistent in each category. BUAP with the worst F1 score for Subtask B (0.45), is still better than the best team in Scenario 1 s2\_end2end for Subtask B  (0.44). Partial annotation or accuracy for Subtask A proves to be critical, reinforcing again that identifying keyphrase boundaries is the most difficult part of the shared task.

Unlike the Scenario 1, in this case the top two teams used classifiers with lexical features (F1: 0.64) as well as neural networks (F1: 0.63). The first team MayoNLP used SVM with rich feature sets like n-grams, lexical features, orthographic features, whereas the second team UKP/EELECTION used used three different neural network approaches and subsequently combined them via majority voting. Both these methods perform quite similarly. However, a CRF based approach and an SVM with simpler feature sets attempted by the two teams LABDA and BUAP are found to be less effective in this scenario.

MayoNLP applied a simple rule based method for synonym-of relation extraction, and Hearst patterns for hyponym-of relation detection.  The rules for synonym-of detection is based on presence of phrases such as \emph{in terms of}, \emph{equivalently}, \emph{which are called} etc in the text between two keyphrases. Interestingly, the RNN based approach of s2\_end2end in Scenario 1 performs better than MayoNLP without using partial annotation of Subtask A.

\begin{table}[t]
\centering
\begin{tabular}{|l|c|}
\hline
\rowcolor{light-gray} 
Teams      &	Overall  \\
\hline
MIT \cite{dernoncourt2017scienceie}	    & {\bf 0.64} \\
s2\_rel \cite{ammar2017scienceie}	& 0.54 \\
NTNU-2 \cite{barik2017scienceie}	& 0.5 \\
LaBDA \cite{suarezpaniagua2017scienceie}	& 0.38 \\
TTI\_COIN\_rel \cite{tomoki2017scienceie}\footnotemark &	0.1 \\
\hline
\textit{upper bound} &   0.84	\\
\textit{random} & 0.04 \\
\hline
\end{tabular}
\caption{F1 scores of teams participating in Scenario 3 and baseline models. Teams participating in Scenario 3 received partial annotation with respect to Subtask A, and Subtask B. Ranking of the teams is based on overall performance measured in Micro F1.} 
\label{tab:scenario3}
\end{table}

\footnotetext{After the end of the evaluation period, team TTI\_COIN\_rel discovered a bug in preprocessing, leading to low results. Their overall result after having corrected for that error is a Macro F1 of 0.48.}
\addtocounter{footnote}{1}

\subsection{Evaluation Scenario 3}
In this scenario, teams need to solve only Subtask C. Partial annotations were provided to the teams for Subtask B and C. 
Five teams participated in this scenario, and F1 scores ranged from 0.1 to 0.64. Please refer to Table \ref{tab:scenario3} for complete result.

Neural network (NN) based models are found to perform better than other methods in this scenario. The best method by MIT uses a convolutional NN (CNN).  The other method uses two phases of NN and found to be reasonably effective (F1: 0.54). 

On the other hand, application of supervised classification with five different classifiers (SVM, decision tree, random forest, multinomial naive Bayes and k-nearest neighbour) using three different feature selection
techniques (chi square, decision tree, and recursive feature elimination) found close accuracy (F1: 0.5) with the top performing ones. 

LaBDA also use a CNN based method. However, the rule based post-processing and argument ordering strategy applied by MIT seemed to give additional advantage as also observed by them. 

However most of the teams in this scenario outperform, all teams from other scenarios (who did not have access to partial information for Subtask B, and C) in relation prediction. This also asserts the significance of accuracy on Subtask A, and B  in order to perform accurately on Subtask C.

\section{Conclusion}

In this paper, we present the setup and discuss participating systems of SemEval 2017 Task 10 on identifying and classifying keyphrases and relations between them from scientific articles, to which 26 systems were submitted. Successful systems vary in their approaches. Most of them use RNNs, often in combination with CRFs as well as CNNs, however the system performing best for evaluation scenario 1 uses an SVM with a well-engineered lexical feature set. Identifying keyphrases is the most challenging subtask, since the dataset contains many long and infrequent keyphrases, and systems relying on remembering them do not perform well.

\section*{Acknowledgments}

We would like to thank Elsevier for supporting this shared task. Special thanks go to Ronald Daniel Jr. for his feedback on the task setup and Pontus Stenetorp for his advice on brat and shared task organisation.

\bibliography{scienceie2017}
\bibliographystyle{acl_natbib}

\end{document}